\documentclass[conference,a4paper]{IEEEtran}
\usepackage{graphicx}
\graphicspath{ {images/} }
\usepackage{amsmath}

\usepackage{amsfonts}
\usepackage{booktabs}
\usepackage{multirow}
\begin{document}
\title{Facial Motion Prior Networks for \\ Facial Expression Recognition}

% \author{\IEEEauthorblockN{Michael Shell}
% \IEEEauthorblockA{School of Electrical and\\Computer Engineering\\
% Georgia Institute of Technology\\
% Atlanta, Georgia 30332--0250\\
% Email: http://www.michaelshell.org/contact.html}
% \and
% \IEEEauthorblockN{Homer Simpson}
% \IEEEauthorblockA{Twentieth Century Fox\\
% Email: homer@thesimpsons.com}
% \and
% \IEEEauthorblockN{James Kirk\\ and Montgomery Scott}
% \IEEEauthorblockA{Starfleet Academy\\
% San Francisco, California 96678-2391\\
% Telephone: (800) 555--1212\\


\author{\IEEEauthorblockN{Yuedong Chen\IEEEauthorrefmark{1},
Jianfeng Wang\IEEEauthorrefmark{2},
Shikai Chen\IEEEauthorrefmark{3},
Zhongchao Shi\IEEEauthorrefmark{2} and
Jianfei Cai\IEEEauthorrefmark{4}}
\IEEEauthorblockA{\IEEEauthorrefmark{1}Nanyang Technological University, Singapore,
\IEEEauthorrefmark{2}AI Lab, Lenovo Research, China,\\
\IEEEauthorrefmark{3}Southeast University, China,
\IEEEauthorrefmark{4}Monash University, Australia\\
donald.chen@ntu.edu.sg,
\{wangjf17, shizc2\}@lenovo.com,
skchen@seu.edu.cn,
jianfei.cai@monash.edu}

% \IEEEauthorblockA{\IEEEauthorrefmark{2}Second Affiliation\\
% Email: \{second, fourth\}@example.com}
% \IEEEauthorblockA{\IEEEauthorrefmark{3}Third Affiliation\\
% Email: third@example.com}
}


\maketitle

\begin{abstract}
%\boldmath
% The abstract goes here.
Deep learning based facial expression recognition (FER) has received a lot of attention in the past few years. Most of the existing deep learning based FER methods do not consider domain knowledge well, which thereby fail to extract representative features. In this work, we propose a novel FER framework, named Facial Motion Prior Networks (FMPN). Particularly, we  introduce an addition branch to generate a facial mask so as to focus on facial muscle moving regions. To guide the facial mask learning, we propose to incorporate prior domain knowledge by using the average differences between neutral faces and the corresponding expressive faces as the training guidance. Extensive experiments on three facial expression benchmark datasets demonstrate the effectiveness of the proposed method, compared with the state-of-the-art approaches.
\end{abstract}


% \fi
\IEEEpeerreviewmaketitle

\section{Introduction}

Facial expression is one of the most important components in daily communications of human beings. It is generated by movements of facial muscles. While different people have different kinds of facial expressions caused by their own expressive styles or personalities, many studies show that there are several types of basic expressions shared by different peoples with different cultural and ethnic background~\cite{fasel2003automatic}. Research on automatic recognition of such basic facial expressions has drawn great attention during the past decades. 

% traditional methods using domain knowledge
Traditional approaches tend to conduct facial expression recognition (FER) by using Gabor Wavelets, sparse coding, etc., where many studies show that subtracting neutral faces from their corresponding expressive faces can help the algorithms to emphasize on the facial moving areas, and significantly improve the expression recognition rate \cite{bazzo2004recognizing}.

% deep learning based method
Convolutional Neural Networks (CNNs) have been widely applied to FER in the recent years. CNNs can achieve good performance by learning powerful high-level features which are better than those conventional hand-crafted features. There are also some other methods proposing to combine global appearance features with local geometry features for FER. Specifically, they feed facial expression recognition networks with not only the original images, but also their related facial landmarks \cite{jung2015joint}, or optical flow \cite{sun2017deep}. \cite{liu2015inspired} extracts more precise facial features by focusing on some specific local parts, inspired by Action Units (AUs) \cite{shao2018deep, shao2018facial}. Recently, a de-expression framework was proposed in \cite{yang2018facial}. They used Generative Adversarial Networks (GANs) to generate neutral faces by learning and filtering the expressive information which is later used for facial expression classification.

% \begin{figure}[t!]
%     \centering
%     \includegraphics[width=0.47\textwidth]{model_overview}
%     \caption{Overview of the proposed framework. An expressive face $I_e$ is fed into a generator to get a facial-motion mask $M$, which is then applied on $I_e$ to get a moving muscle focused face. The masked face and $I_e$ are then fused by a fusion network, followed by a classification network to predict the expression. Note that the learning of $M$ is guided by pseudo ground truth masks, which are the average differences between neutral faces and their corresponding expressive faces for each basic expression. } \label{fig:model_overview}
% %\jf{Change $I_m$ to $M$. Remove $I_x$ and $I_s$.}
% \end{figure}

% our models & contribution
Despite great progress, the existing deep learning based FER methods still have some limitations. 
Firstly, most methods do not incorporate domain knowledge well, so the global features they extract tend to be less discriminative and less representative for FER. Secondly, although some deep learning based methods consider domain knowledge to extract local geometry information by assuming the availability of the full set of landmarks, optical flows or AUs, their assumptions might not be valid since in many cases we might not have all the extra local geometry information and the AU detection task itself is a challenging one. Finally, GANs based method %\cite{yang2018facial}
is also impractical since it requires the neutral and expressive faces of the same person are simultaneously available for all the training data. In addition, the image quality of the generated faces is hard to control, which directly affects the performance of the subsequent expression classifier.  

Therefore, to address these problems, we propose a novel FER framework. Our contribution can be summarized as follows. Firstly, we design a novel end-to-end deep learning framework
named Facial Motion Prior Networks (FMPN) for FER, where we introduce an addition stream to generate a mask to focus on facial muscle moving regions. Secondly, 
we incorporate prior domain knowledge by using the average differences between neutral faces and the corresponding expressive faces as the guidance for the facial motion mask learning.
Finally, our method outperforms current state-of-the-art results
on two laboratory-controlled datasets and one in-the-wild dataset, which demonstrates the effectiveness of the proposed framework.

%Our basic idea is to extract more focused local features by predicting a facial mask that highlights local facial motion regions related to expressions, while we do not need the full set of landmarks, optical flow or AUs. Learning the facial-motion mask without proper guidance cannot achieve a good performance. Thus, inspired by the tradition methods such as \cite{bazzo2004recognizing}, we compute the average differences between neutral faces and their corresponding expressive faces for each basic expression. Then, we use the average differences as the pseudo ground-truth to guide the learning of the facial-motion mask. Fig.~\ref{fig:proposed_model} illustrates the proposed framework. 

% The contributions of this paper are summarized as follows:
% \begin{itemize}
%  \item We propose a novel end-to-end deep learning framework named Facial Motion Prior Networks (FMPN) for FER, where we introduce an addition stream to generate a mask to focus on facial muscle moving regions.
%  \item We incorporate prior domain knowledge by using the average differences between neutral faces and the corresponding expressive faces as the guidance for the facial motion mask learning. Note that unlike \cite{yang2018facial}, which needs a pair of neutral and expressive faces for each training instance, we only need  pairs of neutral and expressive faces for computing the average differences, which can come from another dataset. 
%  \item Experiments on three laboratory-controlled datasets and one in-the-wild dataset demonstrate the effectiveness of the proposed method.
% \end{itemize}

\begin{figure*}[t!]
    \centering
    \includegraphics[width=0.95\textwidth]{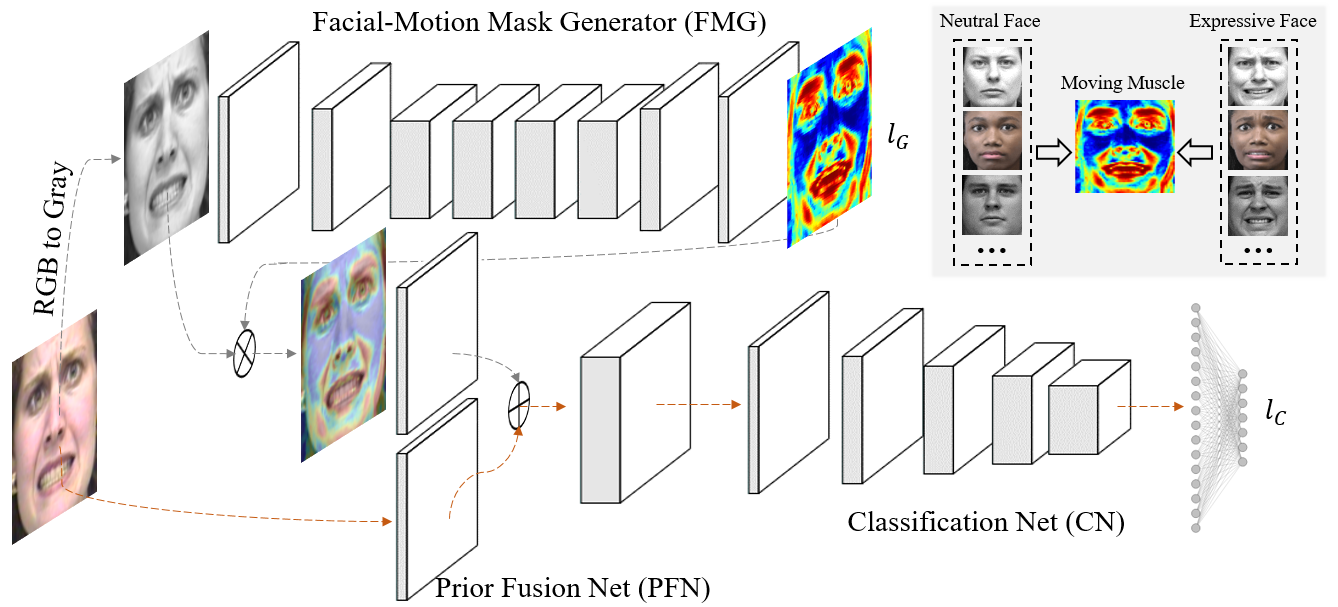}
    \caption{Architecture of the proposed method. The model is composed of three networks: Facial-Motion Mask Generator (FMG), Prior Fusion Net (PFN) and Classification Net (CN). An expressive face is converted to gray scale and fed to FMG to generate a facial-motion mask. Then the mask is applied to and fused with the original input expressive face in PFN. The output of PFN is further fed to CN to extract more powerful features and predict facial expression label. $l_G$ and $l_C$ are loss functions at FMG and CN, respectively, which are end-to-end jointly optimized during training. Note that the learning of FMG is guided by pseudo ground truth masks, which are the average differences between neutral faces and their corresponding expressive faces (see top right corner).} \label{fig:proposed_model}
\end{figure*}

% -------------------------------------------------------------------------
\section{Proposed Approach}
An expressive face of a person is a deformation from the neutral face. In other words, the differences between a neutral face and its corresponding expressive face contain lots of information related to facial expressions. Since basic facial expressions, such as \textit{anger}, \textit{fear}, \textit{happiness} and so on, share traits of uniformity across different people and races, it is reasonable to learn to emphasize specific facial moving parts when conducting expression recognition. On the other hand, emphasizing local facial parts (moving muscle) should not lead to totally ignoring holistic facial image (whole face),  because attributes such as gender and age, provided by the whole face image, can also affect types of expressions significantly. Therefore, both local features and holistic features should be taken into consideration. Inspired by the above analysis, we propose a novel approach to recognize facial expressions.

% \subsection{Overview}
Fig.~\ref{fig:proposed_model} shows the architecture of the proposed FMPN framework, which consists of three networks: \textbf{Facial-Motion Mask Generator (FMG)}, \textbf{Prior Fusion Net (PFN)} and \textbf{Classification Net (CN)}. FMG is constructed to generate a mask, namely facial-motion mask, which highlights moving areas of the given gray scale expressive face. PFN aims to fuse the original input image with the facial-motion mask generated by FMG to introduce domain knowledge to the whole framework. CN is a typical Convolutional Neural Network (CNN) for feature extraction and classification, such as VGG, ResNet or Inception. We mainly discuss FMG and PFN in the following subsections. 

%-------------------------------------------------------------------------
% \begin{figure*}[t!]
%     \centering
%     \begin{subfigure}[b]{0.135\textwidth}
%         \includegraphics[width=\textwidth]{ckp_gt_anger}
%     \end{subfigure}
%     \begin{subfigure}[b]{0.135\textwidth}
%         \includegraphics[width=\textwidth]{ckp_gt_contempt}
%     \end{subfigure}
%     \begin{subfigure}[b]{0.135\textwidth}
%         \includegraphics[width=\textwidth]{ckp_gt_disgust}
%     \end{subfigure}
%     \begin{subfigure}[b]{0.135\textwidth}
%         \includegraphics[width=\textwidth]{ckp_gt_fear.png}
%     \end{subfigure}
%     \begin{subfigure}[b]{0.135\textwidth}
%         \includegraphics[width=\textwidth]{ckp_gt_happy.png}
%     \end{subfigure}
%     \begin{subfigure}[b]{0.135\textwidth}
%         \includegraphics[width=\textwidth]{ckp_gt_sadness.png}
%     \end{subfigure}
%     \begin{subfigure}[b]{0.135\textwidth}
%         \includegraphics[width=\textwidth]{ckp_gt_surprise.png}
%     \end{subfigure}
%     \caption{Ground truth mask of CK+ dataset. The corresponding facial expressions from left to right are: \textit{anger}, \textit{contempt}, \textit{disgust}, \textit{fear}, \textit{happiness}, \textit{sadness} and \textit{surprise}. It can be seen that different facial expressions have different moving muscles, which can benefit the recognition of expressions.}\label{fig:ckp_gt}
% \end{figure*}

% \subsection{Facial-Motion Mask Generator }
\textbf{Facial-Motion Mask Generator (FMG)} is built to generate a facial-motion mask, which is used to highlight expression-related facial motion regions. Instead of making the network to learn active areas blindly, we choose to guide it via some pseudo ground-truth masks that are generated by modeling basic facial expressions. 

In particular, as aforementioned, facial expressions are caused by the contraction of facial muscles, and different people with the same expression share a similar pattern. Therefore, for one specific type of facial expressions, we model muscle moving areas as the difference between an expressive face and its corresponding neutral face, while the characteristic of similarity is modeled by averaging the above differences of all the training instances in the same expression category. Specifically, for a $k$-th type of facial expressions, e.g., \textit{happiness}, its ground truth mask $I_{m}^{(k)}$ is constructed as
\begin{equation}
\label{eq:gen_mask}
I_{m}^{(k)} = \varphi( \frac{1}{N_k}\sum ^{N_k}_i \left | \xi(R^{(k)}_{e,i}) - \xi(R^{(k)}_{n,i}) \right |), 
\end{equation}
where $R^{(k)}_{e}$ is the unprocessed/raw face with the $k$-th type of facial expressions, $R^{(k)}_{n}$ is the corresponding neutral face (faces of the same person shares one neutral face),
%$R_e$ and $R_n$ are the unprocessed/raw images of an face with  and its corresponding neutral face, respectively. 
%Note that $R_e$ and $R_n$ are faces of the same person. 
$N_k$ is the number of faces in the $k$-th expression category, and $\xi(\cdot)$ and $\varphi(\cdot)$ refer to the pre-processing and the post-processing, respectively.

\begin{figure}[t!]
    \centering
    \includegraphics[width=0.47\textwidth]{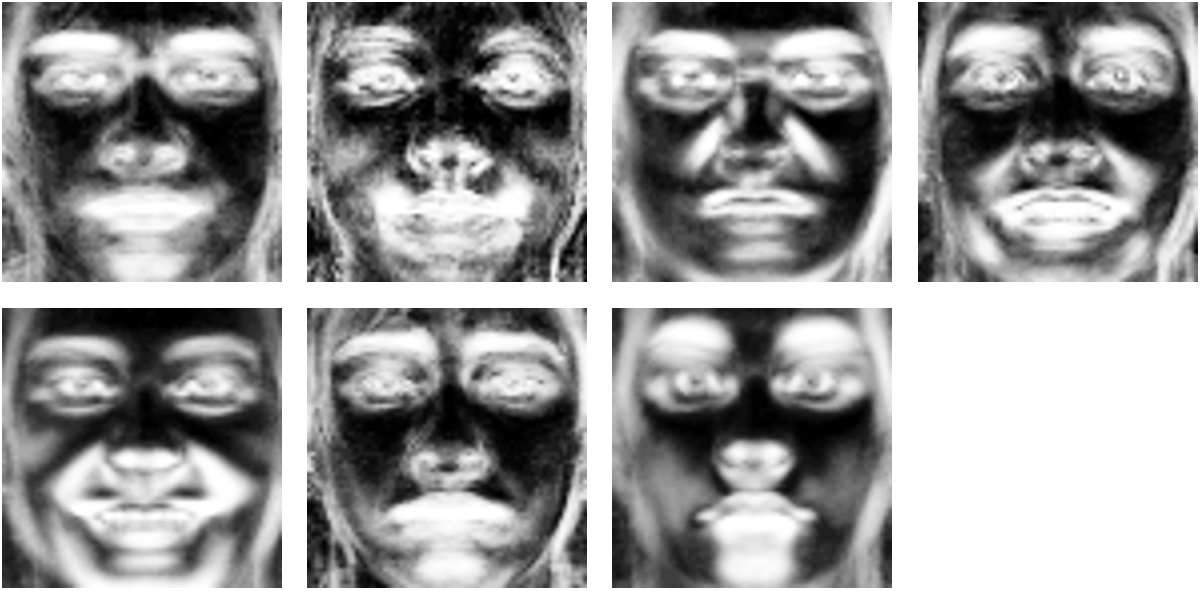}
    \caption{Ground truth masks of the CK+. Corresponding expressions from left to right, top to bottom are: \textit{anger}, \textit{contempt}, \textit{disgust}, \textit{fear}, \textit{happiness}, \textit{sadness} and \textit{surprise}. It can be seen that different facial expressions have different moving muscles, which can benefit the recognition of expressions.}\label{fig:ckp_gt}
\end{figure}

Since facial moving areas are directly modeled as the absolute differences in Eq.~\eqref{eq:gen_mask}, the expressive and neutral faces need to be well aligned, which is ensured by the pre-processing function $\xi(\cdot)$. Specifically, a standard spatial transformation is performed by aligning the detected facial landmarks with the standard reference landmarks. In addition, 
considering that the average face difference is often with close-contrast values, we further introduce a post-processing function $\varphi(\cdot)$ which applies histogram equalization to adjust the difference values for a better distribution. The generated ground truth masks of CK+ \cite{lucey2010extended} for the seven basic expressions are shown in Fig.~\ref{fig:ckp_gt}.

% Network structure design of FMG.
Given an expressive face, FMG is designed to learn a facial-motion mask, trained with the guidance from the pre-computed ground truth mask related to that expression. %Since both input (expressive face) and output (mask) are images in spatial domain, neutral network with fully convolutional layers is a good choice to learn a map between them. 
Particularly, as illustrated in Fig.~\ref{fig:proposed_model}, the input expressive face is first down-sampled and convoluted to extract feature related to geometric structure through two convolutional layers. Then the geometric feature is filtered and transferred to semantic feature related to the dynamic area through four residual blocks. Finally, two transposed convolutional layers are appended to project the learned semantic feature back to spatial domain as moving muscle focused mask. We use MSE (mean square error) for the training objective function of FMG: 
\begin{equation}
\label{eq:loss_mask}
l_G(I_e, k) = \mathbb{E} (f_G(I_e) - I_m^{(k)})^2,
\end{equation}
where $I_e$ is the input expressive face, $I_m^{(k)}$ is the ground truth mask corresponding to the expression class of $I_e$, defined in \eqref{eq:gen_mask}, and $f_G(I_e)$ refers to the mask generated by FMG. 

% Why not just use ground truth mask
One might ask why we do not directly use the computed ground truth masks for facial expression recognition, instead of learning to generate a facial-motion mask. One main reason is that during testing we do not know which ground truth mask should be chosen since different expressions have different facial-motion masks. % In addition, learning to generate a facial-motion mask also helps fine-tune the pre-defined facial-motion regions for better expression recognition. 
We would also like to point out that, considering expressive faces with the same expression in different datasets have similar moving muscles, the ground truth masks obtained from one dataset are mostly likely to be suitable for another dataset. This can help overcome the limitation that some dataset may not contain paired expressive and neutral faces to compute ground truth masks.

% \begin{figure}[t!]
%     % \centering
%     \begin{subfigure}[b]{0.115\textwidth}
%         \includegraphics[width=\textwidth]{ckp_gt_anger}
%     \end{subfigure}
%     \begin{subfigure}[b]{0.115\textwidth}
%         \includegraphics[width=\textwidth]{ckp_gt_contempt}
%     \end{subfigure}
%     \begin{subfigure}[b]{0.115\textwidth}
%         \includegraphics[width=\textwidth]{ckp_gt_disgust}
%     \end{subfigure}
%     \begin{subfigure}[b]{0.115\textwidth}
%         \includegraphics[width=\textwidth]{ckp_gt_fear}
%     \end{subfigure}
%     \vskip\baselineskip
%     \begin{subfigure}[b]{0.115\textwidth}
%         \includegraphics[width=\textwidth]{ckp_gt_happy}
%     \end{subfigure}
%     \begin{subfigure}[b]{0.115\textwidth}
%         \includegraphics[width=\textwidth]{ckp_gt_sadness}
%     \end{subfigure}
%     \begin{subfigure}[b]{0.115\textwidth}
%         \includegraphics[width=\textwidth]{ckp_gt_surprise}
%     \end{subfigure}
%     \caption{Ground truth mask of the CK+. Corresponding expressions from left to right are: \textit{anger}, \textit{contempt}, \textit{disgust}, \textit{fear}, \textit{happiness}, \textit{sadness} and \textit{surprise}.}\label{fig:ckp_gt}
% \end{figure}

% \subsection{Prior Fusion Net}
\textbf{Prior Fusion Net (PFN)} is designed to automatically fuse the original input face with the face that is masked by the facial-motion mask learned from FMG. The former is to provide holistic features, while the latter emphasizes the moving areas, which reflects the common expression definition and domain knowledge. Specifically, PFN produces a fused output $I_s$ by a weighted sum, which can be written as 
\begin{equation}
\label{eq:balance_input}
I_s = w_1 \cdot I_{e^\prime} + w_2 \cdot (I_e \otimes f_G(I_e))), 
\end{equation}
where $I_{e^\prime}$ is the RGB version of the gray-scale image $I_e$, $I_e \otimes f_G(I_e)$ refers to the masked face, which is obtained by element-wise multiplication of face $I_e$ and its corresponding mask $f_G(I_e)$, and $w_1$ and $w_2$ are weights of convolutional layers, which are updated during training.

%\subsection{Total Loss}
% Feed into classifier, Combine training, loss
After PFN, the fused output $I_s$ will then be fed into a CNN based classification network, which can be VGG, ResNet or others. The classification network is trained with the cross entropy loss:
\begin{equation}
\label{eq:loss_cls}
l_C(I_s, k) = - \log(\frac{\exp(f_C(I_s)^{(k)})}{\sum_i^K\exp(f_C(I_s)^{(i)})} ),
\end{equation}
where $f_C(I_s)^{(i)}$ is the $i$-th output value of the classification network $f_C(\cdot)$, $k$ is the target expression, and $K$ is the total number of facial expression classes in a given dataset.
%where $N$ is the total number of facial expressions in a given dataset and $k$ indicates the target expression. $I_s$ is obtained from Eqn. \eqref{eq:balance_input}. And $f_C(\cdot)$ refers to CN.
%
%In general, an input expressive face is read in gray scale and fed to FMG to generate a facial-motion mask, which will be mapped to the input to generate a focused face. And the input face will be read in RGB scale and fused with the above focused face through PFN. The output of PFN will be forwarded to a CNN to extract more features and predict expression. And the 
In this way, the total loss becomes
\begin{equation}
\label{eq:loss_total}
    l_{total} = \lambda_1 \cdot l_G + \lambda_2 \cdot l_C,
\end{equation}
where $l_G$ and $l_C$ are defined in \eqref{eq:loss_mask} and \eqref{eq:loss_cls}, respectively, and $\lambda_1$ and $\lambda_2$ are hyperparameters, being empirically set in training.

\section{Experiments}
% The proposed approach is verified on two laboratory-controlled datasets, including CK+ \cite{lucey2010extended}, MMI \cite{pantic2005web}, and a large scale in-the-wild dataset, AffectNet \cite{mollahosseini2017affectnet}. The settings and results on each dataset are detailed in the following sections.

\subsection{Implementation Details}
% network and baseline setting 
% In our implementation, 
We use Inception V3 as the CN, while other CNN models such as VGG, ResNet can also be adopted. For all datasets, five landmarks %(centers of two eyes, nose and two sides of mouth)
are extracted, followed by face normalization.
% to % the standard reference face with size 
% $96\times112$. Aligned faces are then cropped to $70\times70$, making the nose landmark in the center of the image. 
%To adapt to the input size of Inception, images are resized to $320\times320$ using bilinear interpolation and then cropped to $299\times299$ when fed to the model. On-the-fly data augmentation are employed in training. Specifically, input faces are randomly cropped from four corners or center, and randomly flipped horizontally.  Note that the input faces, including gray and RGB scales, and the ground truth mask must share the exactly same transformation. Training list is shuffled at the beginning of each training epoch. 
In training, input %faces 
are randomly cropped from four corners or center. 
%Specifically, faces are reshaped to $320\times320$ using bilinear interpolation before being cropped to $299\times299$, in order to adapt to the input size of Inception.
Random horizontal flip is also employed. 
%Note that the input faces, including gray and RGB scales, and the ground truth mask must share the same transformation. 
Training list is shuffled at the beginning of each training epoch.

% hyper parameters and training setting 
The CN is initialized using parameters pretrained on ImageNet, while others are randomly initialized. The training starts by tuning only FMG for 300 epochs, using Adam optimizer. The learning rate is initialized as $10^{-4}$ and decayed linearly to 0 from epoch 150. After that, the whole framework is jointly trained, with $\lambda_1 = 10$ and $\lambda_2 = 1$ in \eqref{eq:loss_total}. The learning rate for FMG is reset to $10^{-5}$, while the rest uses $10^{-4}$. We jointly train the entire framework for another 200 epochs, and linearly decay the learning rates from epoch 100. The proposed model is implemented using PyTorch, and code is available at 
\textit{https://github.com/donydchen/FMPN-FER}.
% \url{https://github.com/donydchen/FMPN-FER}. 

% \begin{table}[t!]
% \centering
% \caption{Confusion matrix on the CK+.} 
% \begin{tabular}{c||ccccccc}
% \toprule
%   & AN            & CO            & DI            & FE            & HA           & SA            & SU            \\ 
% \midrule\midrule
% AN & \textbf{95.6} & 0             & 2.2           & 0             & 0            & 2.2           & 0             \\
% CO & 0             & \textbf{98.1} & 0             & 0             & 0            & 0             & 1.9           \\
% DI & 0.6           & 0             & \textbf{99.4} & 0             & 0            & 0             & 0             \\
% FE & 0             & 0             & 0             & \textbf{98.7} & 1.3          & 0             & 0             \\
% HA & 0             & 0             & 0             & 0             & \textbf{100} & 0             & 0             \\
% SA & 2.4           & 0             & 0             & 0             & 0            & \textbf{97.6} & 0             \\
% SU & 0             & 2.0           & 0             & 0             & 0            & 0.8           & \textbf{97.2} \\
% \bottomrule
% \end{tabular}
% \label{tab:ckp_acc_conmat}
% \end{table}
% ------------------- CK+ Acc ------------------------

\subsection{Expression Recognition Results}
% For all datasets, 
We consider two baseline methods: one is using only the classification network, referred as \textit{CNN (baseline)}, the other is using the entire framework but without the training guidance from the ground truth masks, referred as \textit{CNN (no $l_G$)}. 

% CK+
% The \textbf{Extended Cohn-Kanade Dataset (CK+)} \cite{lucey2010extended} is one of the most widely used laboratory-controlled benchmark datasets for FER. %CK+ contains 123 subjects with 593 video sequences. In each sequence, the subject's face changes from neutral status to peak expressive status. Among these subjects, 
% In CK+, 118 subjects with 327 sequences are labelled with seven basic expression labels, namely \textit{anger, contempt, disgust, fear, happiness, sadness} and \textit{surprise}. 
% We conduct our experiment by following the settings of \cite{yang2018facial}. Specifically, the last three frames of each labelled sequence are extracted, which results in a dataset with 981 images. These images are grouped according to the person identity and resort in an ascending order. Then 10-fold cross-validation experiments are conducted, where subjects in training and testing sets are mutually exclusive.

\begin{figure}[t!]
    \centering
    \includegraphics[width=0.47\textwidth]{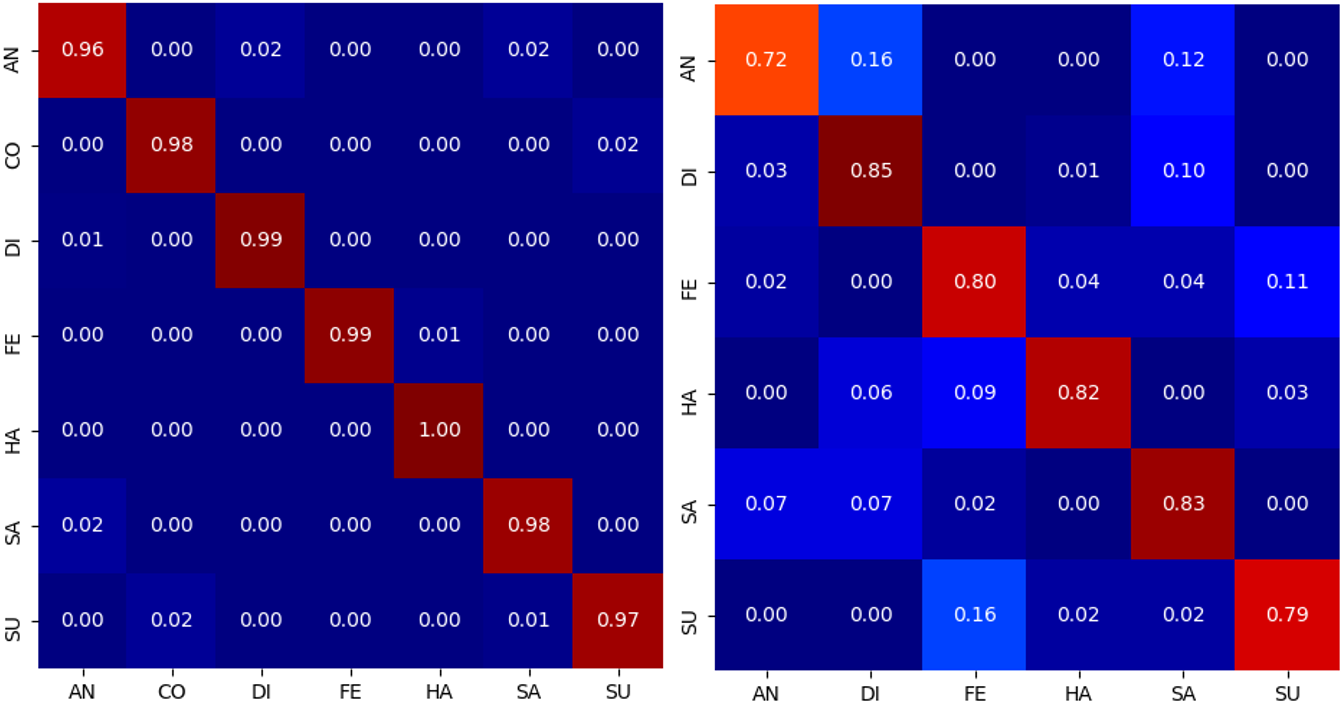}
    \caption{Confusion matrix on the CK+ (left) and MMI (right).} \label{fig:acc_conmat}
\end{figure}

The \textbf{Extended Cohn-Kanade Dataset (CK+)} \cite{lucey2010extended} is a laboratory-controlled benchmark dataset labelled with seven basic expressions. Following the settings of \cite{yang2018facial}, the last three frames of each labelled sequence are extracted, results in a dataset with 981 images. These images are grouped according to the person identity and resort in an ascending order. And 10-fold subject-independent cross-validation experiments are conducted.
% Table~\ref{tab:ckp_acc} shows the quantitative result averaged over 10 runs. 
As shown in TABLE~\ref{tab:ckp_acc}, our proposed method outperforms all other state-of-the-art approaches.%, no matter in sequence-based or image-based settings. 
Compared with the baselines, \textit{CNN(baseline)} and \textit{CNN(no $l_G$)}, our final model achieves a large gain, which indicates that the guidance from the moving muscle mask benefits FER. %Table~\ref{tab:ckp_acc_conmat}
Fig.~\ref{fig:acc_conmat} (left) gives the details of the average accuracy in confusion matrix. We can see that \textit{happiness} is the easiest one to be recognized, likely due to its unique feature of moving muscle around mouth (see Fig.~\ref{fig:ckp_gt}), and \textit{anger} has the relatively lowest recognition rate, which has some confusion with \textit{disgust} and \textit{sadness}.

The \textbf{MMI} \cite{pantic2005web} is another laboratory-controlled dataset labelled with six basic expressions (without \textit{contempt}). As a typical procedure, three peak frames around the center of each labelled sequence are selected, results in a dataset with a total of 624 expressive faces. Similar to CK+, 10-fold person-independent cross-validation experiments are conducted.
% Table~\ref{tab:ckp_acc} reports the average accuracy over 10 runs on MMI.
As illustrated in TABLE~\ref{tab:ckp_acc}, compared with the image-based approaches, our method outperforms them by over 9.51\%. Compared with the sequence-based methods, which use temporal information, our approach still achieves over 7.62\% accuracy improvement. In addition, the large gaps between the two baseline models and our final model further demonstrate the importance of introducing the facial-motion mask and the usefulness of the guidance from the pre-computed ground-truth masks. %Table~\ref{tab:mmi_acc_conmat}
Fig.~\ref{fig:acc_conmat} (right) gives the confusion matrix. We can see that %, \textit{disgust} has the best recognition rate, followed by \textit{happiness}, and  \textit{anger} has the lowest recognition rate, mainly confused with \textit{disgust} and \textit{sadness}, which 
it is similar to those of CK+. This indicates that facial expressions share similar patterns across different datasets.

\begin{table}[!t]
\caption{Average accuracy on the CK+, MMI and AffectNet.}
\label{tab:ckp_acc}
\centering
\begin{tabular}{@{}lcccc@{}}
\toprule
\multirow{2}{*}{Approach} & \multirow{2}{*}{Setting} & \multicolumn{3}{c}{Accuracy (\%)} \\ \cmidrule(l){3-5} 
&  & CK+ & MMI  & AffectNet      \\ \midrule
LBP-TOP~\cite{zhao2007dynamic}      & sequence-based    & 88.99          & 59.51           &   -     \\ 
HOG 3D~\cite{klaser2008spatio}      & sequence-based    & 91.44          & 60.89           &   -     \\ 
DTAGN(Joint)~\cite{jung2015joint}   & sequence-based    & 97.25          & 70.24           &   -     \\ 
STM-Explet~\cite{liu2014learning}   & sequence-based    & 94.19          & 75.12           &   -     \\ 
IACNN~\cite{meng2017identity}       & image-based       & 95.37          & 71.55           &   -     \\ 
DeRF~\cite{yang2018facial}          & image-based       & 97.30          & 73.23           &   -     \\ \midrule
CNN (baseline)                      & image-based       & 90.78          & 68.81           & 60.86   \\
CNN (no $l_G$)                      & image-based       & 91.82          & 63.10           & 60.01   \\ 
\textbf{FMPN (Ours)}                & image-based       & \textbf{98.06} & \textbf{82.74}  & \textbf{61.52} \\ \bottomrule
\end{tabular}
\end{table}

% ------------------- CK+ & MMI con mat ------------------
% \begin{figure}[t!]
%     \centering
%     \begin{subfigure}[b]{0.235\textwidth}
%         \includegraphics[width=\textwidth]{ckp_acc_conmat}
%     \end{subfigure}
%     \begin{subfigure}[b]{0.235\textwidth}
%         \includegraphics[width=\textwidth]{mmi_acc_conmat}
%     \end{subfigure}
%     \caption{Confusion matrix on the CK+ (left) and MMI (right).}
%     \label{fig:acc_conmat}
% \end{figure}

% AffectNet
The \textbf{AffectNet} \cite{mollahosseini2017affectnet} is a very large in-the-wild dataset.
% an in-the-wild dataset contains more than 1 million faces, collected by querying three major search engines using expression related keywords in 6 languages. Among them, nearly half of the images are manually annotated with facial expression labels. 
We conduct experiments on a subset of AffectNet. % to further verify our proposed approach. 
% In particular, only faces with seven basic expression labels are taken into consideration. 
We randomly select around 3500 images for each of the seven basic expressions, resulting in a total of 24530 images. We randomly split it into training and testing set with a ratio of 9 to 1. Since the neutral faces corresponding to expressive faces are not available, we cannot generate the ground truth masks based on this dataset itself. 
%However, because faces are labeled with expression labels based on the Facial Action Coding System (FACS) \cite{ekman2002facial}, 
%However, there are similarities shared among different datasets, as the definition of basic expressions and the results of CK+, MMI shown. 
Thus, we borrow the ground truth masks from CK+. %to train our framework with the constructed subset of AffectNet. Since AffectNet has enough data, we directly train the entire framework jointly for 200 epochs, removing the pretraining for FMG which we did in laboratory-controlled datasets. All other settings, including hyperparameters, learning rates, etc., remain unchanged.
Since AffectNet has enough data, we remove the pretraining for FMG. %which we did in laboratory-controlled datasets. 
All other settings, including hyperparameters, learning rates, etc., remain unchanged.

% The accuracy of AffectNet is reported  
As shown in TABLE~\ref{tab:ckp_acc}, transferring muscle moving masks from CK+ improves the recognition rate, which suggests that the information of facial moving muscles can be shared across different datasets, and it does help improve the performance of expression recognition. Note that it seems that the gain is not as significant as those in other datasets. This is mainly because of the large number of test images in AffectNet and the challenge of dealing with in-the-wild images. Other state-of-the-art methods did not report their results on AffectNet. 

% \begin{table}[t!]
% \centering
% \begin{tabular}{lr}
% \toprule
% Approach             & Accuracy      \\ \midrule
% CNN (baseline)       & 0.609          \\ 
% CNN (no $l_G$)       & 0.600          \\ \midrule
% \begin{tabular}[l]{@{}l@{}}\textbf{FMPN (Ours)}\\ (with average masks from CK+)\end{tabular} & \textbf{0.615} \\ \midrule
% \end{tabular}
% \caption{Average accuracy comparison on the AffectNet dataset with seven basic expressions.}
% \label{tab:affectnet_acc}
% \end{table}

% \begin{figure}[!t]
% \centering
% \includegraphics[width=0.15\textwidth]{ckp_acc_conmat}
% \caption{Simulation Results}
% \label{fig_sim}
% \end{figure}

% \begin{figure}[!t]
% \centering
% \includegraphics[width=0.15\textwidth]{mmi_acc_conmat}
% \caption{Simulation Results}
% \label{fig_sim}
% \end{figure}

% -------------------------------------------------------------------------
\section{Conclusion}
We have proposed a novel FER framework, which incorporates prior knowledge. % of facial expressions into FER. 
Particularly, for a given expressive face, we generate a facial mask to focus on facial muscle moving regions and we use the average differences between neutral faces and expressive faces as the guidance for the facial mask learning. Our method achieves the best results on CK+ and MMI datasets % and the second best results on Oulu-CASIA. 
We have also reported our results on the large-scale in-the-wild dataset, AffectNet. % These outstanding results show the effectiveness of the proposed facial motion mask learning and its usefulness in FER. 
These outstanding results demonstrates the effectiveness of the proposed model in FER.


\section*{Acknowledgment}
This research is partially supported by NTU BeingTogether Centre and Monash FIT Start-up Grant.




\bibliographystyle{IEEEtran}
\bibliography{IEEEabrv,references}


% \end{thebibliography}

% \nocite{*}

\end{document}